\documentclass[11pt]{article}

\usepackage[final]{acl}

\usepackage{times}
\usepackage{latexsym}
\usepackage[T1]{fontenc}
\usepackage[utf8]{inputenc}
\usepackage{microtype}
\usepackage{inconsolata}
\usepackage{graphicx}

\usepackage{subcaption}
\usepackage{amsmath}
\usepackage{booktabs}
\usepackage{xspace}
\usepackage{multirow}
\usepackage{pifont}
\usepackage{tcolorbox}
\usepackage{amssymb}
\usepackage{threeparttable}
\usepackage{mathtools}
\usepackage{amsthm}
\usepackage{bm}
\usepackage{algorithm}
\usepackage{algorithmic}
\usepackage{colortbl}
\usepackage[capitalize,noabbrev]{cleveref}
\usepackage{placeins}

\theoremstyle{plain}

\theoremstyle{definition}

\theoremstyle{remark}

\newcommand{\sys}{\textsc{ShardMemo}\xspace}

\title{ShardMemo: Scope-Before-Routing for Agentic Memory Retrieval}

\author{
  \textbf{Yang Zhao}$^{1,*,\dagger}$ \quad
  \textbf{Chengxiao Dai}$^{2,*}$ \quad
  \textbf{Mengying Kou}$^{3}$ \\
  \textbf{Yue Xiu}$^{4}$ \quad
  \textbf{Dusit Niyato}$^{3}$ \\
  $^{1}$Independent Researcher \quad
  $^{2}$The University of Sydney \\
  $^{3}$Nanyang Technological University \\
  $^{4}$University of Electronic Science and Technology of China \\
  $^{*}$Equal contribution. \quad
  $^{\dagger}$Corresponding author: \texttt{S180049@e.ntu.edu.sg}
}

\begin{document}
\maketitle


\begin{abstract}
Agentic systems accumulate persistent memory across sessions, tools,
and tasks, and a later request must retrieve from it under two distinct
constraints: which memories it is permitted to access, and which are
relevant under a limited search budget. Existing memory systems handle
access scope in two flawed ways: applying scope \emph{after}
retrieval wastes probe budget on inadmissible memories, while treating
scope as a learned ranking feature makes a hard constraint depend on
router quality. We present \textsc{ShardMemo}, an agentic memory system
built on \emph{scope-before-routing}: metadata predicates first
identify the admissible shards, and a learned router then selects a
small number of them for shard-local approximate nearest neighbor
retrieval. Given the supplied scope predicate and metadata, this separates
hard admissibility from learned relevance ranking, so inadmissible shards
cannot consume shard-probe budget. We
evaluate on LoCoMo, HotpotQA, and ToolBench, covering conversational,
long-context, and procedural memory. Under matched supervision and
fixed budgets, \textsc{ShardMemo} improves over a learned router
baseline by roughly $+3$ F1 on LoCoMo; in end-to-end LoCoMo evaluation
it improves over the strongest evaluated memory baseline by up to
$+6.8$ F1, with gains on HotpotQA and ToolBench.
\end{abstract}

\section{Introduction}
\label{sec:intro}

Agentic systems built on large language models (LLMs) increasingly rely
on persistent memory across sessions, tools, and tasks. These memories
include dialogue history, user preferences, tool outputs, task
trajectories, and reusable procedures; later requests may depend on
information recorded weeks earlier or in another
session~\cite{packer2023memgpt,park2023generativeagents,maharana2024locomo,wang2023voyager,qian2026toward}.
We refer to retrieval over these persistent stores as \emph{agentic
memory retrieval}.

Agentic memory retrieval is not ordinary semantic retrieval. A memory
may be relevant to the current task but inadmissible under the
request's user, session, time, tool namespace, or permission state.
Relevance can be learned from retrieval signals, whereas admissibility
is metadata-defined and must be enforced as a hard constraint.

This creates a retrieval-ordering problem. Memory systems that layer
admissibility onto a semantic retriever apply scope only \emph{after}
retrieval---as a post-filter on returned evidence, so limited search
budget may already have been spent on memories that were never
admissible. The alternative, treating scope as one feature among others
in a learned ranker, makes a hard access constraint depend on router
quality: a routing error can return inadmissible memory. Neither is
acceptable when scope is a hard constraint and budget is fixed. Our
focus is how an agentic memory system should allocate a small retrieval
budget when access scope must be enforced relative to a supplied policy.

We introduce \sys, an agentic memory system centered on
\emph{scope-before-routing}. Metadata predicates first construct the
admissible shard set, and a learned masked router then selects a small
number of admissible shards for shard-local approximate nearest
neighbor (ANN) retrieval. Unlike predicate-filtered ANN, which enforces
constraints inside a single index, \sys composes hard predicate masking
with a budget-constrained router over heterogeneous shards, so
inadmissible shards never consume probe budget while the router still
learns which admissible shards are most likely to contain useful
evidence.

\sys applies this principle to three memory roles common in agentic
workflows: active context for interpreting the current request,
declarative memory for long-horizon evidence retrieval, and procedural
memory for reusable tool-use skills. We evaluate the same scoped
retrieval interface on LoCoMo, HotpotQA, and ToolBench, covering
long-horizon conversational memory, long-context multi-hop retrieval,
and procedural skill reuse~\cite{maharana2024locomo,yang-etal-2018-hotpotqa,qin2023toolllm}.
We test whether enforcing admissibility before routing improves
retrieval under fixed budgets, against prior memory systems and
controlled routing baselines.

\textbf{Contributions.}
This paper makes three contributions:
\begin{itemize}
  \item We formulate agentic memory retrieval as scope-constrained,
  budgeted shard selection.
  \item We introduce \sys, a scope-before-routing memory system
  composing hard predicate masking with a budget-constrained router
  over admissible shards.
  \item We evaluate the same scoped retrieval interface, showing through controlled routing, noise, scaling, and ablation analyses where scope-before-routing improves the quality--efficiency frontier.
\end{itemize}

\section{Background and Related Work}
\label{sec:related}

\textbf{Agentic memory.}
Retrieval-augmented generation grounds language models in external evidence
\cite{lewis2020rag,karpukhin2020dpr}. Agentic memory systems
extend this to persistent state across interactions: MemGPT separates working
context from external memory \cite{packer2023memgpt}, Generative Agents store
profiles and episodic records \cite{park2023generativeagents}, and Reflexion
reuses prior trajectories through self-reflection \cite{shinn2023reflexion}.
Recent systems such as A-Mem, LightMem, Mem0, MemoryOS, and GAM improve how
memories are organized, summarized, consolidated, or reused
\cite{amem2025agentmemory,lightmem2025,mem02025,memoryos2025,gam2025,
wang2026memguard,liu2026conmem,zhang2026memmark}.

\textbf{Heterogeneous memory for agents.}
Agentic memory is not a single uniform store. Stable user facts, turn-level
observations, multi-turn summaries, working context, tool outputs, and reusable
tool-use procedures serve different roles and carry different access patterns;
tool-using agents in particular benefit from reusing procedures and workflows
\cite{yao2023react,schick2023toolformer,wang2023voyager,
zhang2024agentworkflowmemory,liu2026palm}.
\sys reflects this heterogeneity with bounded active context, sharded
declarative evidence, and a separate procedural skill library, rather than
forcing all memory types into one retrieval layout.

\textbf{Constrained and budgeted retrieval.}
ANN systems make vector search efficient
\cite{johnson2019faiss,guo2020scann,malkov2016hnsw,subramanya2019diskann};
filtered ANN improves how to search an index under metadata predicates
\cite{gollapudi2023filtereddiskann,jin2024curator}, and federated search
studies how to select collections under resource limits
\cite{callan2000cori,shokouhi2011federated}. Agentic memory adds a further
demand: request-level admissibility over persistent, heterogeneous stores,
where a semantically relevant memory may be inaccessible because of user,
session, time, tool namespace, or permission state. \sys constructs the
admissible shard set from metadata first, then performs relevance- and
cost-aware routing only within that set, and remains compatible with filtered
ANN inside each shard. This addresses a question filtered ANN and federated
search do not: how an agent should allocate a small per-request probe budget
across admissible memory shards.

\section{Method}
\label{sec:method}

\subsection{Overview}
\label{sec:overview}

\sys is an agentic memory system for scope-constrained, budgeted retrieval. It
implements \emph{scope-before-routing}: metadata-defined admissibility first
determines which memory shards a request may access, and learned routing then
allocates probe budget only within that admissible set. \sys instantiates this
principle through Tier~A active context, Tier~B sharded declarative evidence
retrieval, and Tier~C reusable skill retrieval with fallback to Tier~B.

\begin{figure*}[htbp]
  \centering
  \includegraphics[width=0.95\linewidth]{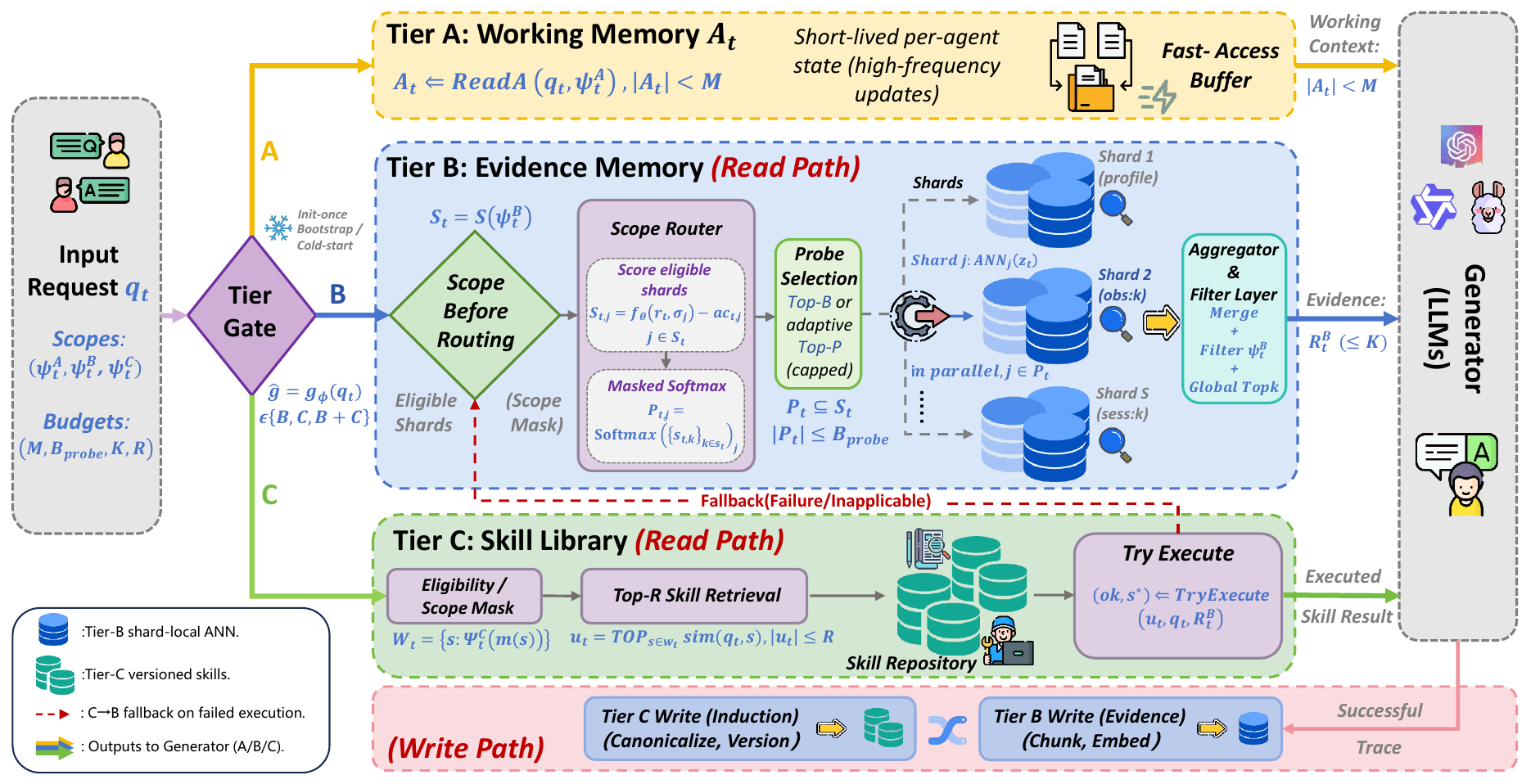}
  \caption{\sys read path. Tier~A supplies active context, a tier gate activates
  Tier~B and/or Tier~C, and Tier~B applies the eligibility mask before routing.
  Tier~C falls back to Tier~B when no executable skill applies.}
  \label{fig:shardmemo}
\end{figure*}

\subsection{Problem Formulation}
\label{sec:scope_budget}

We formalize each memory read as budgeted retrieval under hard scope
constraints. Let $q$ be a request. The system may return Tier~A active context
$A(q)$, Tier~B evidence $R^B(q)$, and Tier~C skills $\mathcal{U}(q)$; let
$\mathcal{P}(q)$ denote the probed Tier~B shard set. Each tier
$\tau\in\{A,B,C\}$ defines a metadata predicate
\[
\psi^\tau(m,q)\in\{0,1\},
\]
which determines whether metadata $m$ is admissible for request $q$. For a
request sequence, we write $q_t$ for the request at time $t$ and
$\psi_t^\tau(m)=\psi^\tau(m,q_t)$.

For Tier~B, let $S$ be the number of shards. Each shard
$j\in\{1,\ldots,S\}$ has metadata $m_j$, inducing the admissible shard set
\[
\mathcal{S}_B(q)=\{j:\psi^B(m_j,q)=1\}.
\]
Under budgets $(M,B_{\mathrm{probe}},K,R)$, which cap Tier~A context, probed
Tier~B shards, returned Tier~B evidence, and retrieved Tier~C skills, the read
path must satisfy
\begin{align*}
|A(q)| &\le M,
&
|\mathcal{P}(q)| &\le B_{\mathrm{probe}},
\\
|R^B(q)| &\le K,
&
|\mathcal{U}(q)| &\le R,
\\
\mathcal{P}(q) &\subseteq \mathcal{S}_B(q).
\end{align*}
The router ranks within the admissible set rather than deciding which shards are allowed, so routing errors can reduce recall within that set but cannot spend probe budget outside the request's scope.

\subsection{Tier~A: Active Context}
\label{sec:tiera}

Tier~A stores bounded active context: session metadata, recent turns,
plans, and tool outputs. Its role is query interpretation, not large-scale
retrieval. For example, Tier~A can resolve temporal or context-dependent
references such as ``last week'' or ``that plan'' before Tier~B searches
long-term evidence:
\[
A_t=\mathrm{ReadA}(q_t,\psi_t^A), \qquad |A_t|\le M.
\]

\subsection{Tier~B: Scope-Before-Routing for Evidence Retrieval}
\label{sec:tierb}

Tier~B stores declarative evidence in $S$ shards, each with a shard-local ANN
index $\mathrm{ANN}_j$. Shards are organized into profile shards for stable
entity facts, observation shards for turn-level records, and session shards for
multi-turn summaries. This layout separates evidence by granularity and access
cost.

\textbf{Eligibility masking.}
For request $q_t$, the admissible shard set is
\[
\mathcal{S}_t=\mathcal{S}_B(q_t)\subseteq\{1,\ldots,S\}.
\]
All shards outside $\mathcal{S}_t$ are masked to logit $-\infty$ before
routing, so they cannot be selected or consume probe budget.

\textbf{Budgeted routing and retrieval.}
Let $\bm{z}_t=\mathrm{Embed}(q_t)$ be the query embedding and
$\bm{r}_t=[\bm{z}_t;\bm{\varphi}_t]$ be the router input, where
$\bm{\varphi}_t$ contains request-time metadata features such as session, time,
tool namespace, permission state, and budget flags. Within $\mathcal{S}_t$, the
learned shard scorer $f_\theta$ scores each admissible shard:
\[
s_{t,j}=f_\theta(\bm{r}_t,\bm{\sigma}_j)-\alpha c_{t,j},
\qquad j\in\mathcal{S}_t,
\]
where $\bm{\sigma}_j$ is a shard summary, $\alpha$ is the cost-bias coefficient,
and $c_{t,j}=n_j/\bar n$ is a latency proxy based on the number of vectors
$n_j$ in shard $j$ and the mean shard size $\bar n$. Router probabilities are
computed only over admissible shards:
\[
p_{t,j}=\mathrm{softmax}\big(\{s_{t,k}\}_{k\in\mathcal{S}_t}\big)_j .
\]
The router selects $\mathcal{P}_t\subseteq\mathcal{S}_t$ with
$|\mathcal{P}_t|\le B_{\mathrm{probe}}$ using Top-$B_{\mathrm{probe}}$; we also
evaluate an adaptive Top-$P$ efficiency variant capped by
$B_{\mathrm{probe}}$.

Let $L$ be the per-shard ANN candidate cap. Candidates from probed
shards are merged, checked again at the item level, and globally ranked:
\begin{align*}
\widetilde{\mathcal C}_t
  &= \bigcup_{j\in\mathcal P_t} \mathrm{ANN}_j(\bm z_t,L), \\
\mathcal C_t
  &= \{x\in\widetilde{\mathcal C}_t:\psi_t^B(m(x))=1\}, \\
R_t^B
  &= \operatorname{TopK}_{x\in\mathcal C_t}^{K}
     \operatorname{score}(\bm z_t,x).
\end{align*}

\textbf{Router training.}
When evidence-to-shard supervision is available, each query $q_t$ is paired
with a gold shard set $G_t\subseteq\mathcal{S}_t$ containing admissible shards
that store supporting evidence.

We optimize a multi-positive routing loss:
\[
L_{\mathrm{route}}(t)=-\log\sum_{j\in G_t}p_{t,j},
\]
which rewards probability mass on any supporting shard. To align training with
fixed-budget inference, we also use
\[
L_{\mathrm{route}}^{\mathrm{BA}}(t)
= -\log\!\sum_{j\in G_t} w_{t,j}p_{t,j}.
\]
Here, $w_{t,j}\in[0,1]$ softly emphasizes gold shards near the $B_{\mathrm{probe}}$ probe window. When evidence-to-shard supervision is unavailable, \sys uses no-label routing, such as cosine similarity to shard summaries, as an ablation. Algorithm~\ref{alg:tierBread} gives the Tier-B read procedure.

\subsection{Tier~C: Skill Memory}
\label{sec:tierc}

\textbf{Read.} Tier C filters by scope, keeps the latest in-scope version
of each skill, and retrieves up to $R$ skills:
\begin{align*}
\mathcal W_t
  &= \{s\in\mathcal W:\psi_t^C(m(s))=1\}, \\
\mathcal W_t^{\mathrm{cur}}
  &=
  \Bigl\{s\in\mathcal W_t:
  \mathrm{ver}(s)=
  \max_{\substack{s'\in\mathcal W_t\\ \mathrm{id}(s')=\mathrm{id}(s)}}
  \mathrm{ver}(s')\Bigr\}, \\
\mathcal U_t
  &=
  \operatorname{TopR}^{R}_{s\in\mathcal W_t^{\mathrm{cur}}}
  \operatorname{sim}(q_t,s).
\end{align*}
The executor orders $\mathcal U_t$ by reliability and similarity, fills
slots from $q_t$ and $\mathcal E_t$, validates preconditions, executes,
and validates the result. If no skill succeeds, ShardMemo falls back to
Tier B with the original Tier-B budget.

\textbf{Write.}
Tier~C induces skills from successful tool-use traces. Given a trace,
\sys canonicalizes the procedure, abstracts constants into typed slots,
checks whether the canonicalized procedure matches an existing skill, and
either reuses that skill identifier or creates a new one. It then synthesizes
deterministic validation tests and stores the skill only if those tests pass.
New successful traces for an existing procedure append a new versioned record
rather than overwriting earlier validated behavior. This keeps procedural reuse
separate from Tier~B evidence retrieval while using the same scope-predicate
interface. Appendix~\ref{app:tierc_algs} gives the read and write algorithms.


\section{Experiments}
\label{sec:experiments}

\subsection{Setup}
\label{sec:exp_setup}

\textbf{Benchmarks.}
We evaluate \sys on three public benchmarks. LoCoMo
\cite{maharana2024locomo} tests long-horizon conversational memory across
single-hop, multi-hop, temporal, and open-domain questions; we exclude
adversarial questions because they primarily test instruction-following rather
than memory retrieval. HotpotQA \cite{yang-etal-2018-hotpotqa} tests
long-context multi-hop evidence retrieval under the MemAgent protocol
\cite{yu2025memagentreshapinglongcontextllm}, using 56K/224K/448K-token inputs
constructed from gold documents plus distractors. ToolBench
\cite{qin2023toolllm} tests procedural skill reuse with in-domain and
family-level out-of-domain splits. 

\begin{figure}[!ht]
\centering
\includegraphics[width=\linewidth]{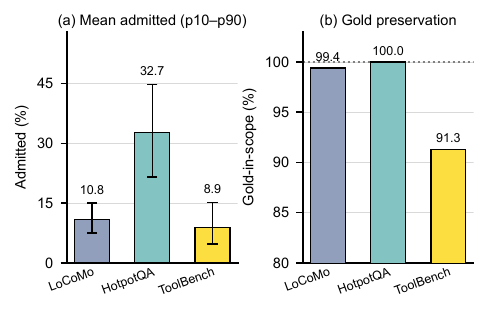}
\caption{Admissible-set statistics under the scope predicates. Filtered
unit: Tier-B shards (LoCoMo/HotpotQA), Tier-C skills (ToolBench).
(a) Mean admitted \% per request with p10--p90 whiskers.
(b) Gold-in-scope \%.}
\label{fig:scope_stats}
\end{figure}

\noindent\textbf{Scope predicates.}
Scope-before-routing matters when the admissible shard or skill set is a
request-dependent subset of the memory store. On LoCoMo, the Tier-B predicate
$\psi^B$ admits shards matching the conversation and speaker(s) referenced by
the request. On HotpotQA, $\psi^B$ admits only document shards that appear in
the request's input context, filtering out the global distractor pool. On
ToolBench, the Tier-C predicate $\psi^C$ admits skills whose tool namespace is
authorized for the request. These predicates are strict rather than no-ops:
Figure~\ref{fig:scope_stats} reports the fraction of memory units admitted per
request and the fraction of queries for which the gold evidence or executable
skill remains in scope. ToolBench gold-in-scope falls below 100\% because some
requests have no executable skill in the authorized namespace, motivating the
Tier-C $\to$ Tier-B fallback in Section~\ref{sec:tierc}.

\noindent\textbf{Metrics.}
We report token-level F1 and BLEU-1 on LoCoMo, answer F1 on HotpotQA, and
Precision@$R$ plus StepRed on ToolBench. StepRed is the mean reduction in
tool-use steps after a valid skill is reused. For retrieval efficiency, VecScan
is the mean number of vectors scanned per query across the probed shards, and
p95 latency is measured from scope filtering through candidate ranking,
excluding generator decoding. ShardHit@$B_{\mathrm{probe}}$ measures whether at
least one probed shard contains gold evidence; we use it as a routing diagnostic
rather than a full multi-hop evidence-recall metric. Full formal definitions of all metrics (ShardHit@$B_{\mathrm{probe}}$, Precision@$R$, StepRed, VecScan, and p95 latency) are given in Appendix~\ref{app:metrics}.

\noindent\textbf{Baselines.}
For end-to-end memory retrieval, we compare against Vanilla LLM, centralized
RAG, and five agent-memory systems: A-Mem \cite{amem2025agentmemory},
LightMem \cite{lightmem2025}, Mem0 \cite{mem02025}, MemoryOS
\cite{memoryos2025}, and GAM \cite{gam2025}. For controlled Tier-B routing,
we compare cosine-to-prototype routing, an untrained router, a simple learned
router trained with the same evidence-to-shard supervision, and component
ablations. For Tier-C skill retrieval, we compare recency, BM25, embedding
similarity, trace kNN, and a static skill library.

\noindent\textbf{Comparison protocol.}
Within each benchmark, we match prompts, output post-processing, and evidence
or skill budgets where applicable. Retrieval baselines use the same evidence
items and chunking policy. Centralized RAG uses the same embedding model and
ANN configuration to search a single global index over the evidence store.
Controlled Tier-B routing baselines additionally use the same shard layout,
shard-local indexes, and per-request budgets as \sys, isolating the routing
policy; methods without scope-aware routing enforce admissibility only by
post-filtering returned evidence, which tests retrieval order under fixed
shard-probe budgets rather than serving as a production filtered-vector-search
benchmark. The LoCoMo routing comparisons use evidence-to-shard supervision
where stated, so they should be read as supervised routing comparisons under
matched budgets, not as training-free comparisons.

\noindent\textbf{Implementation and splits.}
We use \texttt{all-MiniLM-L6-v2} embeddings and FAISS-HNSW indexes. Tier-B
retrieval is shard-local; centralized RAG uses a single global index. In \sys,
final evidence ranking combines BM25, dense similarity, and route score with
weights $0.4/0.4/0.2$. The router is a two-layer MLP trained with Adam.
Unless otherwise stated, the answer generator is GPT-OSS-120B with
temperature 0, and all results are mean over three random seeds with
seed standard deviation shown in blue subscript.
LoCoMo uses an 8/2 conversation-level split (1{,}226/314 QA pairs); ToolBench
uses a 70/30 induction/evaluation split plus a family-level OOD split. We run
all experiments on a single server without inter-request concurrency. Full
hyperparameters, hardware, software, and dataset statistics are in
Appendix~\ref{app:exp_details}.

\subsection{Main Results}
\label{sec:results}

\begin{table*}[t]
\centering
\caption{Results on LoCoMo with the GPT-OSS-120B backbone.}
\label{tab:locomo_results}
\small
\setlength{\tabcolsep}{3pt}
\renewcommand{\arraystretch}{1.15}
\begin{tabular}{lcccccccc}
\toprule
\rowcolor{teal!10}
\multicolumn{9}{c}{\textbf{LoCoMo}} \\
\midrule
\multirow{2}{*}{Method}
& \multicolumn{2}{c}{Single-Hop}
& \multicolumn{2}{c}{Multi-Hop}
& \multicolumn{2}{c}{Temporal}
& \multicolumn{2}{c}{Open Domain} \\
\cmidrule(lr){2-3}\cmidrule(lr){4-5}\cmidrule(lr){6-7}\cmidrule(lr){8-9}
& F1 & BLEU-1 & F1 & BLEU-1 & F1 & BLEU-1 & F1 & BLEU-1 \\
\midrule
Vanilla LLM
  & $44.72_{\textcolor{blue}{\pm0.11}}$ & $35.28_{\textcolor{blue}{\pm0.17}}$
  & $27.10_{\textcolor{blue}{\pm0.21}}$ & $21.45_{\textcolor{blue}{\pm0.09}}$
  & $22.86_{\textcolor{blue}{\pm0.14}}$ & $17.02_{\textcolor{blue}{\pm0.22}}$
  & $14.25_{\textcolor{blue}{\pm0.11}}$ & $11.86_{\textcolor{blue}{\pm0.16}}$ \\
\rowcolor{gray!6}
RAG
  & $51.36_{\textcolor{blue}{\pm0.28}}$ & $45.80_{\textcolor{blue}{\pm0.31}}$
  & $28.33_{\textcolor{blue}{\pm0.17}}$ & $21.63_{\textcolor{blue}{\pm0.26}}$
  & $43.57_{\textcolor{blue}{\pm0.23}}$ & $37.26_{\textcolor{blue}{\pm0.34}}$
  & $21.54_{\textcolor{blue}{\pm0.19}}$ & $15.29_{\textcolor{blue}{\pm0.21}}$ \\
\midrule
A-MEM
  & $44.38_{\textcolor{blue}{\pm0.37}}$ & $36.42_{\textcolor{blue}{\pm0.42}}$
  & $26.81_{\textcolor{blue}{\pm0.24}}$ & $20.45_{\textcolor{blue}{\pm0.33}}$
  & $46.20_{\textcolor{blue}{\pm0.29}}$ & $37.14_{\textcolor{blue}{\pm0.46}}$
  & $11.86_{\textcolor{blue}{\pm0.31}}$ & $11.10_{\textcolor{blue}{\pm0.27}}$ \\
\rowcolor{gray!6}
LIGHTMEM
  & $42.30_{\textcolor{blue}{\pm0.26}}$ & $37.77_{\textcolor{blue}{\pm0.39}}$
  & $30.28_{\textcolor{blue}{\pm0.35}}$ & $25.49_{\textcolor{blue}{\pm0.28}}$
  & $41.48_{\textcolor{blue}{\pm0.41}}$ & $36.82_{\textcolor{blue}{\pm0.24}}$
  & $17.13_{\textcolor{blue}{\pm0.22}}$ & $13.81_{\textcolor{blue}{\pm0.36}}$ \\
MEM0
  & $45.35_{\textcolor{blue}{\pm0.44}}$ & $37.95_{\textcolor{blue}{\pm0.33}}$
  & $38.60_{\textcolor{blue}{\pm0.27}}$ & $26.54_{\textcolor{blue}{\pm0.48}}$
  & $47.64_{\textcolor{blue}{\pm0.36}}$ & $41.27_{\textcolor{blue}{\pm0.31}}$
  & $28.32_{\textcolor{blue}{\pm0.29}}$ & $20.68_{\textcolor{blue}{\pm0.41}}$ \\
\rowcolor{gray!6}
MEMORYOS
  & $47.17_{\textcolor{blue}{\pm0.32}}$ & $42.39_{\textcolor{blue}{\pm0.25}}$
  & $34.13_{\textcolor{blue}{\pm0.39}}$ & $26.02_{\textcolor{blue}{\pm0.30}}$
  & $42.40_{\textcolor{blue}{\pm0.22}}$ & $32.81_{\textcolor{blue}{\pm0.37}}$
  & $19.83_{\textcolor{blue}{\pm0.27}}$ & $14.40_{\textcolor{blue}{\pm0.34}}$ \\
GAM
  & $58.38_{\textcolor{blue}{\pm0.41}}$ & $53.00_{\textcolor{blue}{\pm0.35}}$
  & $41.17_{\textcolor{blue}{\pm0.26}}$ & $33.06_{\textcolor{blue}{\pm0.44}}$
  & $59.52_{\textcolor{blue}{\pm0.33}}$ & $52.70_{\textcolor{blue}{\pm0.29}}$
  & $34.10_{\textcolor{blue}{\pm0.24}}$ & $26.85_{\textcolor{blue}{\pm0.38}}$ \\
\midrule
\textbf{ShardMemo}
  & $\mathbf{64.08}_{\textcolor{blue}{\pm0.23}}$ & $\mathbf{59.70}_{\textcolor{blue}{\pm0.30}}$
  & $\mathbf{46.28}_{\textcolor{blue}{\pm0.27}}$ & $\mathbf{39.65}_{\textcolor{blue}{\pm0.21}}$
  & $\mathbf{66.34}_{\textcolor{blue}{\pm0.19}}$ & $\mathbf{60.31}_{\textcolor{blue}{\pm0.32}}$
  & $\mathbf{40.13}_{\textcolor{blue}{\pm0.17}}$ & $\mathbf{33.82}_{\textcolor{blue}{\pm0.25}}$ \\
\bottomrule
\end{tabular}
\end{table*}

\noindent \textbf{LoCoMo.}
Table~\ref{tab:locomo_results} reports the broad benchmark comparison on long-horizon conversational memory. \sys achieves the best F1 and BLEU-1 in all four non-adversarial categories. Compared with GAM, the strongest evaluated memory baseline, \sys improves F1 by +5.70 on Single-Hop, +5.11 on Multi-Hop, +6.82 on Temporal, and +6.03 on Open Domain. The Temporal gains are largest, matching the design role of Tier~A active context and scoped Tier~B retrieval.

In addition, LoCoMo gains are not tied to the 120B generator. With GPT-OSS-20B and
Qwen3-32B backbones, \sys improves over GAM by 6.76 and 6.34 average F1, respectively. Temporal gains are the most consistent across backbones (+11.75 and +11.50 F1), supporting the role of Tier~A deictic anchoring and scoped Tier~B retrieval. Full per-category backbone results are in Appendix~\ref{app:backbone} (Table~\ref{tab:locomo_results_extra}).

\begin{table}[t]
\centering
\caption{Results on HotpotQA with the GPT-OSS-120B backbone.}
\label{tab:hotpotqa_results}
\small
\setlength{\tabcolsep}{4pt}
\renewcommand{\arraystretch}{1.15}
\begin{tabular}{lccc}
\toprule
\rowcolor{teal!10}
\multicolumn{4}{c}{\textbf{HotpotQA}} \\
\midrule
\multirow{2}{*}{Method}
& \multicolumn{3}{c}{F1} \\
\cmidrule(lr){2-4}
& 56K & 224K & 448K \\
\midrule
Vanilla LLM
  & $54.43_{\textcolor{blue}{\pm0.14}}$
  & $52.10_{\textcolor{blue}{\pm0.19}}$
  & $49.08_{\textcolor{blue}{\pm0.13}}$ \\
\rowcolor{gray!6}
RAG
  & $52.08_{\textcolor{blue}{\pm0.25}}$
  & $50.75_{\textcolor{blue}{\pm0.22}}$
  & $51.34_{\textcolor{blue}{\pm0.28}}$ \\
\midrule
A-MEM
  & $34.83_{\textcolor{blue}{\pm0.36}}$
  & $31.06_{\textcolor{blue}{\pm0.42}}$
  & $29.77_{\textcolor{blue}{\pm0.31}}$ \\
\rowcolor{gray!6}
LIGHTMEM
  & $37.68_{\textcolor{blue}{\pm0.33}}$
  & $35.42_{\textcolor{blue}{\pm0.41}}$
  & $29.82_{\textcolor{blue}{\pm0.37}}$ \\
MEM0
  & $30.13_{\textcolor{blue}{\pm0.39}}$
  & $28.40_{\textcolor{blue}{\pm0.45}}$
  & $29.52_{\textcolor{blue}{\pm0.34}}$ \\
\rowcolor{gray!6}
MEMORYOS
  & $28.82_{\textcolor{blue}{\pm0.37}}$
  & $23.56_{\textcolor{blue}{\pm0.32}}$
  & $22.18_{\textcolor{blue}{\pm0.43}}$ \\
GAM
  & $62.10_{\textcolor{blue}{\pm0.34}}$
  & $60.92_{\textcolor{blue}{\pm0.29}}$
  & $57.40_{\textcolor{blue}{\pm0.38}}$ \\
\midrule
\textbf{ShardMemo}
  & $\mathbf{64.37}_{\textcolor{blue}{\pm0.21}}$
  & $\mathbf{62.65}_{\textcolor{blue}{\pm0.23}}$
  & $\mathbf{58.79}_{\textcolor{blue}{\pm0.26}}$ \\
\bottomrule
\end{tabular}
\end{table}

\noindent \textbf{HotpotQA.}
Table~\ref{tab:hotpotqa_results} evaluates the same scope-before-routing
interface for long-context multi-hop evidence retrieval. \sys reaches
64.37/62.65/58.79 F1 at 56K/224K/448K tokens, improving over GAM by
+2.27/+1.73/+1.39. These gains are smaller than on LoCoMo, which is expected:
as context budgets grow, the marginal benefit of precise shard selection
decreases. We therefore treat HotpotQA as evidence that scoped
routing transfers to long-context evidence retrieval.

\begin{table}[t]
\centering
\caption{Tier-C skill retrieval on ToolBench ($R=3$) with the GPT-OSS-120B
backbone.}
\label{tab:toolbench_tierc_results}
\small
\setlength{\tabcolsep}{10pt}
\renewcommand{\arraystretch}{1.15}
\begin{threeparttable}
\begin{tabular}{lcc}
\toprule
\rowcolor{teal!10}
\multicolumn{3}{c}{\textbf{ToolBench}} \\
\midrule
Method & P@$R$\,$\uparrow$ & StepRed\,$\uparrow$ \\
\midrule
\rowcolor{gray!6}
Recency
  & $0.79_{\textcolor{blue}{\pm0.02}}$
  & $1.44_{\textcolor{blue}{\pm0.07}}$ \\
BM25
  & $0.84_{\textcolor{blue}{\pm0.02}}$
  & $1.70_{\textcolor{blue}{\pm0.08}}$ \\
\rowcolor{gray!6}
Embedding similarity
  & $0.88_{\textcolor{blue}{\pm0.02}}$
  & $1.79_{\textcolor{blue}{\pm0.07}}$ \\
Trace kNN
  & $0.69_{\textcolor{blue}{\pm0.04}}$
  & $1.65_{\textcolor{blue}{\pm0.11}}$ \\
\rowcolor{gray!6}
Static skill library
  & $0.86_{\textcolor{blue}{\pm0.02}}$
  & $1.76_{\textcolor{blue}{\pm0.08}}$ \\
\midrule
\textbf{Tier-C Skills}
  & $\mathbf{0.97}_{\textcolor{blue}{\pm0.01}}$
  & $\mathbf{2.01}_{\textcolor{blue}{\pm0.04}}$ \\
\bottomrule
\end{tabular}
\end{threeparttable}
\end{table}

\noindent \textbf{ToolBench.}
Table~\ref{tab:toolbench_tierc_results} evaluates the procedural-memory path.
Tier~C achieves the best Precision@3 (0.97) and StepRed (2.01). Compared with
the strongest baseline, Tier~C improves Precision@3 by +0.09 and StepRed by
+0.22. The StepRed gain indicates that retrieved skills are more often
executable end-to-end, not merely similar in embedding space.

\subsection{Quality--Efficiency Trade-off}
\label{sec:tradeoff}

\begin{figure}[t]
  \centering
  \includegraphics[width=\linewidth]{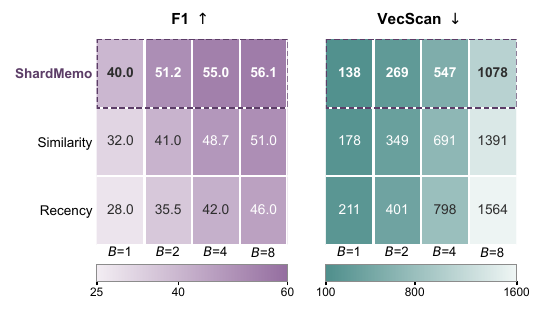}
  \caption{F1 (left, $\uparrow$) and VecScan (right, $\downarrow$) on
  LoCoMo across $B_{\mathrm{probe}}\in\{1,2,4,8\}$ at $K{=}10$.
  Darker is better.}
  \label{fig:pareto_locom}
\end{figure}

We test with a controlled routing comparison (Table~\ref{tab:ablation}, router-baselines block) and an iso-budget sweep in Figure~\ref{fig:pareto_locom}.

\textbf{Controlled routing comparison.}
The router-baselines block of Table~\ref{tab:ablation} compares routing policies
under an identical shard layout and supervision where applicable. A simple
learned router improves over cosine routing, showing that evidence-to-shard
supervision is useful. The full model further improves F1 while reducing VecScan
and p95 latency, indicating that the full routing stack improves the
quality--efficiency operating point beyond a simpler learned scorer.

\textbf{Iso-budget sweep.}
Figure~\ref{fig:pareto_locom} sweeps the shard-probe budget
$B_{\mathrm{probe}}\in\{1,2,4,8\}$ at fixed $K=10$. We use powers-of-two probe
budgets to show the trend across small and large retrieval budgets; the main
operating point, $B_{\mathrm{probe}}=3$, is reported in
Table~\ref{tab:ablation}. Across budgets, \sys achieves higher F1 at comparable
or lower VecScan than similarity and recency routing. Centralized retrieval can
recover evidence by searching broadly, but its cost grows sharply.

\textbf{Scope before vs.\ after routing.}
Enforcing scope \emph{before} routing, rather than post-filtering, is the source
of both gains. Post-filtering returns the same admissible evidence but spends
probe budget on inadmissible shards, lowering F1 from $54.21$ to $51.68$ by
reducing within-scope recall and raising VecScan from $414$ to $560$
(Table~\ref{tab:ablation}).

\begin{table}[t]
\centering
\caption{Shard-count scaling at $B_{\mathrm{probe}}{=}3$, $K{=}10$.
$^\dagger$ = default.}
\label{tab:shard_scalability}
\small
\setlength{\tabcolsep}{6pt}
\renewcommand{\arraystretch}{1.15}
\begin{tabular}{c c c c c c}
\toprule
\rowcolor{teal!10}
\multicolumn{6}{c}{\textbf{Shard-count scaling}} \\
\midrule
$S$ & F1 & ShardHit & VecScan & p95 & Route(ms) \\
\midrule
5  & 51.5 & \textbf{0.93} & 580 & 86 & \textbf{0.2} \\
\rowcolor{gray!6}
10 & 52.8 & 0.89 & 485 & 78 & 0.4 \\
20 & 53.7 & 0.85 & 438 & \textbf{74} & 0.7 \\
\rowcolor{gray!6}
$40^\dagger$ & \textbf{54.2} & 0.82 & 414 & 76 & 1.2 \\
80 & 53.8 & 0.76 & \textbf{388} & 81 & 2.2 \\
\bottomrule
\end{tabular}
\end{table}

\begin{table}[t]
\centering
\caption{Label-noise robustness ($B_{\mathrm{probe}}{=}3$, $K{=}10$);
cosine $=$ no-supervision floor.}
\label{tab:label_noise}
\small
\setlength{\tabcolsep}{5pt}
\renewcommand{\arraystretch}{1.15}
\begin{tabular}{l c c c c}
\toprule
\rowcolor{teal!10}
\multicolumn{5}{c}{\textbf{Label-noise robustness}} \\
\midrule
Noise $\epsilon$ & F1 & ShardHit & VecScan & p95 \\
\midrule
0\% (clean)  & 54.21 & 0.82 & 414 & 76 \\
\rowcolor{gray!6}
20\%         & 52.14 & 0.79 & 431 & 79 \\
40\%         & 49.87 & 0.73 & 462 & 85 \\
\midrule
\rowcolor{gray!6}
Cosine (no labels) & 47.34 & 0.67 & 521 & 95 \\
\bottomrule
\end{tabular}
\end{table}

\subsection{Component Ablations}
\label{sec:ablation}

\begin{table*}[t]
\centering
\caption{Ablations on LoCoMo ($B_{\mathrm{probe}}{=}3$, $K{=}10$).}
\label{tab:ablation}
\small
\setlength{\tabcolsep}{8pt}
\renewcommand{\arraystretch}{1}
\begin{threeparttable}
\begin{tabular}{p{0.48\textwidth} c c c c}
\toprule
\textbf{Variant}
& \textbf{F1}{\scriptsize$\uparrow$}
& \textbf{ShardHit}{\scriptsize$\uparrow$}
& \textbf{VecScan}{\scriptsize$\downarrow$}
& \textbf{p95}{\scriptsize$\downarrow$} \\
\midrule
\rowcolor{teal!10}
\multicolumn{5}{c}{\textbf{\textit{Router baselines}}} \\
\midrule
Cosine similarity to shard prototype
& $47.34_{\textcolor{blue}{\pm0.22}}$ & $0.67_{\textcolor{blue}{\pm0.01}}$ & $521_{\textcolor{blue}{\pm7}}$ & $95_{\textcolor{blue}{\pm2}}$ \\
\rowcolor{gray!6}
Simple learned router (same supervision)
& $51.14_{\textcolor{blue}{\pm0.31}}$ & $0.75_{\textcolor{blue}{\pm0.01}}$ & $472_{\textcolor{blue}{\pm8}}$ & $86_{\textcolor{blue}{\pm2}}$ \\
Untrained router (no supervision)
& $39.82_{\textcolor{blue}{\pm0.48}}$ & $0.52_{\textcolor{blue}{\pm0.02}}$ & $627_{\textcolor{blue}{\pm16}}$ & $116_{\textcolor{blue}{\pm5}}$ \\
\midrule
\rowcolor{teal!10}
\multicolumn{5}{c}{\textbf{\textit{Scope-first, no learned router}}} \\
\midrule
Scoped Global ANN
& $53.74_{\textcolor{blue}{\pm0.19}}$ & --- & $782_{\textcolor{blue}{\pm27}}$ & $113_{\textcolor{blue}{\pm6}}$ \\
\rowcolor{gray!6}
Scoped All-Admissible (equal $K$, probe all admissible)
& $54.58_{\textcolor{blue}{\pm0.17}}$ & --- & $1065_{\textcolor{blue}{\pm38}}$ & $142_{\textcolor{blue}{\pm8}}$ \\
\midrule
\rowcolor{teal!10}
\multicolumn{5}{c}{\textbf{\textit{Component ablations}}} \\
\midrule
w/o cost-aware gating ($\alpha{=}0$)
& $53.47_{\textcolor{blue}{\pm0.29}}$ & $0.80_{\textcolor{blue}{\pm0.01}}$ & $478_{\textcolor{blue}{\pm12}}$ & $86_{\textcolor{blue}{\pm3}}$ \\
\rowcolor{gray!6}
w/o eligibility masking (post-filter only)
& $51.68_{\textcolor{blue}{\pm0.37}}$ & $0.73_{\textcolor{blue}{\pm0.02}}$ & $560_{\textcolor{blue}{\pm21}}$ & $100_{\textcolor{blue}{\pm5}}$ \\
w/o adaptive Top-$P$ (fixed Top-$B_{\mathrm{probe}}$)
& $53.85_{\textcolor{blue}{\pm0.24}}$ & $0.81_{\textcolor{blue}{\pm0.01}}$ & $465_{\textcolor{blue}{\pm11}}$ & $85_{\textcolor{blue}{\pm3}}$ \\
\rowcolor{gray!6}
w/o heterogeneous shard families (session only)
& $51.12_{\textcolor{blue}{\pm0.41}}$ & $0.77_{\textcolor{blue}{\pm0.02}}$ & $601_{\textcolor{blue}{\pm24}}$ & $106_{\textcolor{blue}{\pm6}}$ \\
w/o Tier~A active context
& $50.38_{\textcolor{blue}{\pm0.34}}$ & $0.73_{\textcolor{blue}{\pm0.02}}$ & $509_{\textcolor{blue}{\pm17}}$ & $93_{\textcolor{blue}{\pm4}}$ \\
\rowcolor{gray!6}
w/o budget-aware surrogate ($L_{\mathrm{route}}$ only)
& $53.07_{\textcolor{blue}{\pm0.32}}$ & $0.78_{\textcolor{blue}{\pm0.01}}$ & $428_{\textcolor{blue}{\pm10}}$ & $78_{\textcolor{blue}{\pm3}}$ \\
\midrule
\textbf{ShardMemo (full)}
& $\mathbf{54.21}_{\textcolor{blue}{\pm0.24}}$ & $\mathbf{0.82}_{\textcolor{blue}{\pm0.01}}$ & $\mathbf{414}_{\textcolor{blue}{\pm9}}$ & $\mathbf{76}_{\textcolor{blue}{\pm2}}$ \\
\bottomrule
\end{tabular}
\begin{tablenotes}[flushleft]\footnotesize
\item \textit{Note:} ShardHit is ShardHit@$B_{\mathrm{probe}}$; it is undefined (---) for scope-first baselines that do not select a bounded shard set. VecScan is the mean number of vectors scanned per query. p95 is the 95th-percentile retrieval latency (ms).
\end{tablenotes}
\end{threeparttable}
\end{table*}

Having established the routing mechanism in Section~\ref{sec:tradeoff}, we now
isolate the contribution of each remaining component on LoCoMo under fixed
budgets (Table~\ref{tab:ablation}). Heterogeneous
shard families matter: collapsing Tier~B to session-only shards lowers F1 to
51.12 and raises VecScan to 601. Removing Tier~A active context lowers F1 to
50.38, consistent with its role in temporal and deictic grounding. Cost-aware
gating and adaptive Top-$P$ mainly affect efficiency: disabling them changes F1
modestly but increases VecScan and p95 latency.

\subsection{Robustness and Scaling}
\label{sec:robust_scaling}

All results use LoCoMo with
$B_{\mathrm{probe}}=3$ and $K=10$ unless otherwise stated.

\textbf{Shard-count scaling.}
Table~\ref{tab:shard_scalability} varies the number of Tier-B shards. F1 remains
within 2.7 points across $S\in\{5,10,20,40,80\}$ and peaks at $S=40$, our
default setting. VecScan does not grow with the total number of shards; it
remains controlled by the fixed probe budget and the size of the probed shards.
Routing overhead increases with $S$ but stays below 3\,ms at $S=80$. The drop in
ShardHit at $S=80$ suggests that overly fine shard maps can fragment supporting
evidence.

\begin{figure}[t]
\centering
\includegraphics[width=\linewidth]{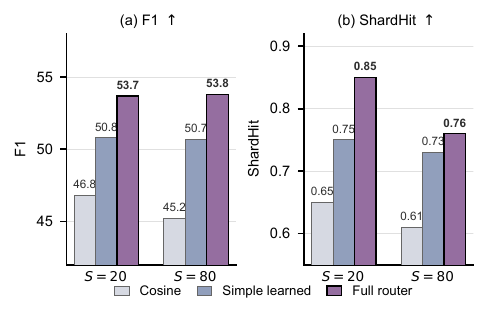}
\caption{Router comparison across shard counts ($B_{\mathrm{probe}}=3$,
$K=10$). (a) F1 and (b) ShardHit for cosine, simple learned, and full
routers at $S\in\{20,80\}$.}
\label{fig:router_shards}
\end{figure}

\textbf{Router comparison across scales.}
Figure~\ref{fig:router_shards} crosses router type with shard count. The full
router maintains its advantage over the simple learned router at both $S=20$ and
$S=80$, indicating that the family-aware routing stack does not degrade relative
to simpler routers as the shard set grows.

\textbf{Label-noise robustness.}
To test whether routing depends on clean shard labels, we corrupt only the
training supervision by replacing each gold shard in $G_t$ with a uniformly
sampled incorrect admissible shard with probability $\epsilon$. Test labels
remain clean, and the router is retrained from scratch at each noise level.
Table~\ref{tab:label_noise} shows graceful degradation: even at 40\% label
noise, the trained router remains above the no-label cosine baseline.

\subsection{Case Study}
\label{sec:case_study}

\begin{figure}[htbp]
\centering
\footnotesize
\begin{tcolorbox}[
  colback=teal!3!white,
  colframe=teal!55!black,
  title={\small Case Study: Tier-C Skill Retrieval with C$\to$B Fallback},
  rounded corners,
  width=\linewidth,
  boxsep=3pt, left=6pt, right=6pt, top=5pt, bottom=5pt
]
\textbf{Task:} Get contact details of the ACT customs agency in New
Caledonia (ToolBench G1 \#10). \\
\textbf{Tier gate:} C (procedural)
\tcblower
\textbf{Tier~C retrieval} ($R{=}3$, scope: \texttt{transitaires}): \\
\texttt{[skill:transitaire\_lookup\_v2]} \\
\texttt{transitaire\_for\_transitaires(is\_id=\{id\})}
\hfill\textit{(0.91 sim.)} \\
\textit{Slot filling:}
\texttt{id=``ACT\_AGENCE\_\allowbreak{}CALEDONIENNE\_\allowbreak{}DE\_TRANSIT''}
(canonicalized via Tier-A directory cache).

\smallskip
\textbf{Execution:} returns \texttt{\{codePostal: ``98800'',
email: ``sales@act.nc'', phone: ``+687 27.55.48''\}}. \\
\textcolor{teal!55!black}{\checkmark\ Success --- StepRed: 2 steps saved
(list + lookup folded into one skill call).}

\vskip 0.9em
\rule{\linewidth}{0.4pt}
\vskip 0.7em

\textbf{Task:} List freight forwarders serving Wallis-and-Futuna. \\
\textbf{Tier gate:} C (procedural)

\smallskip
\textbf{Tier~C retrieval} ($R{=}3$, scope: \texttt{transitaires}): \\
\textit{Max sim.\ $0.34 < \tau_C{=}0.5$ (territory mismatch).} \\
\textcolor{red!60!black}{\textbf{Execution: Failed (inapplicable).}}

\smallskip
\textbf{C$\to$B fallback:} Tier-B activated under same budgets
($B_{\mathrm{probe}}{=}3$, $K{=}10$). \\
\texttt{[doc:transitaires/spec]} ``Covers customs agencies in
New Caledonia (country code NC) only.'' \\
\mbox{}\hfill\textit{(API-spec shard)}

\smallskip
\textbf{Prediction:} The transitaires API does not cover
Wallis-and-Futuna; no freight forwarders are available via this
namespace. \\
\mbox{}\hfill\textcolor{teal!55!black}{\checkmark\ Correct via fallback}
\end{tcolorbox}
\caption{Two Tier-C paths under shared budget
$(B_{\mathrm{probe}}{=}3, K{=}10, R{=}3)$: direct skill execution
(top) and C$\to$B fallback when no skill applies (bottom).}
\label{fig:case_tierc}
\end{figure}

Figure~\ref{fig:case_tierc} illustrates the two Tier-C control paths on
a real ToolBench trace (G1 \#10) and a constructed out-of-scope query.
\textbf{Top:} A customs-lookup request is gated to Tier C; \sys retrieves
a previously induced \texttt{transitaire\_lookup} skill, canonicalizes
the entity ``ACT'' via the Tier-A directory cache, validates the filled
call against the skill's stored tests, and executes against the real
ToolBench endpoint---folding what would otherwise be a \emph{list +
lookup} plan into a single skill call.
\textbf{Bottom:} For a territory outside the API's coverage, no skill
clears $\tau_C$, so \sys activates Tier B under the \emph{same} budget
and recovers the answer from the tool's specification shard; a Tier-C
miss never degrades below the Tier-B baseline. This discipline is what
lets Tier C dominate Table~\ref{tab:toolbench_tierc_results} without
sacrificing recall on out-of-scope requests.
Appendix~\ref{app:case} gives additional Tier-B case studies.

\section{Conclusion}

We presented \sys, a memory architecture for scope-constrained, budgeted
agentic retrieval. The central idea is scope-before-routing: metadata
predicates first mask inadmissible shards, and a learned router then allocates
a small shard-probe budget only among admissible shards. Across conversational
QA, long-context retrieval, and tool-use benchmarks, \sys improves the
quality--efficiency trade-off of memory retrieval under fixed budgets. The
controlled ablations show that eligibility masking is central: replacing
scope-before-routing with post-filtering lowers F1 and increases VecScan.
Separate tiers for active context, declarative evidence, and procedural skills
make the retrieval layer usable across different agent-memory roles without
forcing all memory types into a single layout. Future work should evaluate
dynamic permission changes, shard-map adaptation, concurrent multi-agent
access, and distributed deployment.


\section*{Limitations}

  First, our evaluation is on English-only benchmarks; effectiveness under
  multilingual or cross-lingual workloads is untested.
  Second, Tier~B router training requires evidence$\to$shard annotations;
  without such supervision, the untrained router is significantly weaker
  (39.82 vs.\ 54.21 F1). The no-label cosine variant (47.34 F1) recovers
  roughly half this gap and offers a practical fallback.
  Third, experiments use a fixed shard map; workload drift is described
  architecturally but not evaluated empirically.
  Fourth, all experiments are conducted on a single server with no
  inter-request concurrency; multi-node deployment and concurrent
  multi-agent workloads are out of scope.
  Fifth, ToolBench evaluation uses held-out tool families but does not
  simulate continuous schema evolution; Tier~C gains may depend on relative
  schema stability within the evaluation period.
  Sixth, our shard-count study covers $S\in\{5,10,20,40,80\}$; distributed
  sharding at production scale ($S\gtrsim 10\mathrm{K}$) is not evaluated.
  Seventh, scope enforcement is conditional on externally supplied predicates
  and metadata. False exclusions cap recall; false inclusions waste budget and,
  when item metadata are also incorrect, can violate the intended policy
  despite item-level rechecking. Revocation, policy-conflict resolution,
  propagation, and audit logging remain external to \sys.

\section*{Ethics Statement}

  All experiments use publicly released benchmarks (LoCoMo, HotpotQA,
  ToolBench); LoCoMo contains synthetic personas rather than data about
  real individuals. \sys itself does not collect or disclose personal
  data; its scope predicates are designed to compose with whatever
  access-control policies a deployment specifies. Operators should audit
  shard supervision and admissibility under their own data governance
  before production use.

\section*{Use of AI assistants.}
AI assistants were used to support language editing, LaTeX polishing, checklist
preparation, and clarity checks. They were not used to generate experimental
results, fabricate data, run evaluations, or make final scientific claims. The
authors reviewed and verified all manuscript content.

\bibliography{references}

%
\newpage
\appendix

\section{Notation Table}
\label{app:notation}

\begin{table}[!ht]
\centering
\footnotesize
\setlength{\tabcolsep}{9pt}
\renewcommand{\arraystretch}{1.15}
\begin{tabular}{@{}l p{0.60\linewidth}@{}}
\toprule
\textbf{Symbol} & \textbf{Meaning} \\
\midrule
$q, q_t$ & Request / query (at time $t$). \\
$x, s$ & Tier~B evidence item; Tier~C skill artifact. \\
$m(\cdot)$ & Metadata (scope keys, permissions, provenance). \\
$\psi^\tau(m,q)$ & Tier-$\tau$ scope predicate. \\
$M, B_{\mathrm{probe}}$ & Tier~A and probe budgets. \\
$K, R$ & Tier~B evidence and Tier~C skill budgets. \\
$A_t$ & Tier~A active context. \\
$R^B_t, \mathcal{U}_t$ & Tier~B evidence; Tier~C retrieved skills. \\
$\mathcal{W}, \mathcal{W}_t, \mathcal{W}_t^{\mathrm{cur}}$ & Skill store; admissible skills; latest admissible skill versions. \\
$S,\,j$ & Number of shards; shard index. \\
$\mathrm{ANN}_j$ & Shard-local ANN index. \\
$\mathcal{S}_t$ & Eligible/admissible shard set. \\
$\mathcal{P}_t$ & Probed shard set. \\
$\bm{z}_t, \bm{r}_t$ & Query embedding; router input. \\
$\bm{\varphi}_t$ & Request-time metadata features used by the router. \\
$\bm{\sigma}_j$ & Shard summary. \\
$f_\theta$ & Router scorer. \\
$s_{t,j}, p_{t,j}$ & Shard logit; routing probability. \\
$c_{t,j}, \alpha$ & Cost estimate; cost-bias weight. \\
$\gamma, P_{\min}, P_{\max}$ & Adaptive Top-$P$ parameters. \\
$\tau_t$ & Adaptive Top-$P$ threshold. \\
$G_t$ & Gold shard set. \\
$L_{\mathrm{route}}$ & Set-likelihood routing loss. \\
$L_{\mathrm{route}}^{\mathrm{BA}}$ & Budget-aware routing loss. \\
$w_{t,j}$ & Budget-aware gold weight. \\
$\mathrm{VecScan}$ & Mean vectors scanned per query. \\
$L$ & Per-shard ANN candidate cap. \\
$\mathcal{E}_t$ & Available context for Tier-C slot filling. \\
$\rho(s)$ & Tier-C skill reliability score. \\
$\mathrm{id}(s),\ \mathrm{ver}(s)$ & Skill identifier and version of skill record $s$. \\
\bottomrule
\end{tabular}
\caption{Notation summary.}
\label{tab:notation}
\end{table}

\FloatBarrier

\section{Algorithms}
\label{app:algorithms}

\providecommand{\algfont}{\footnotesize}

Let $[S]=\{1,\ldots,S\}$ denote the set of Tier-B shard indices.
For request $q_t$, write $\psi_t^\tau(m)=\psi^\tau(m,q_t)$.
The admissible Tier-B shard set is
\[
\mathcal{S}_t
=
\mathcal{S}_B(q_t)
=
\{j\in[S]:\psi_t^B(m_j)=1\},
\]
where $m_j$ is the metadata of shard $j$. The probed shard set
$\mathcal{P}_t$ must satisfy
\[
\mathcal{P}_t\subseteq\mathcal{S}_t,
\qquad
|\mathcal{P}_t|\le B_{\mathrm{probe}}.
\]
$\mathrm{ANN}_j(\bm z_t,L)$ returns at most $L$ candidates from shard
$j$. $\operatorname{TopK}^{K}$ and $\operatorname{TopR}^{R}$ return at
most $K$ evidence items and $R$ skills, respectively. For a skill record
$s$, $\mathrm{id}(s)$, $\mathrm{ver}(s)$, and $\rho(s)$ denote its
identifier, version, and reliability score.

\subsection{Tier~B Read Algorithm}
\label{app:tierb_alg}

Algorithm~\ref{alg:tierBread} implements scope-before-routing in three stages:
it first restricts the eligible shard set to those whose metadata satisfy
$\psi_t^B$, then ranks the eligible shards by the cost-corrected router score
$s_{t,j}=f_\theta(\bm r_t,\bm\sigma_j)-\alpha c_{t,j}$ (optionally truncating
the probe list by adaptive Top-$P$ when the routing distribution is sharp),
and finally retrieves shard-local ANN candidates from the probed shards
before re-applying admissibility and a global Top-$K$ re-rank. Returning early
on $\mathcal S_t=\varnothing$ is what gives scope filtering its compute-saving
effect: when no shard is admissible, neither the router nor any ANN index is
invoked.

\begin{algorithm}[t]
\caption{Tier~B read: scope-before-routing and global Top-$K$}
\label{alg:tierBread}
\algfont
\begin{algorithmic}[1]
\REQUIRE $\bm z_t,\bm r_t,\psi_t^B,(B_{\mathrm{probe}},K,L)$,
         $\{\bm\sigma_j,m_j,c_{t,j},\mathrm{ANN}_j\}_{j=1}^{S}$
\ENSURE Evidence $R_t^B$, probed shards $\mathcal P_t$

\STATE $\mathcal S_t\gets\{j\in[S]:\psi_t^B(m_j)=1\}$
\IF{$\mathcal S_t=\varnothing$}
  \STATE \textbf{return} $(\varnothing,\varnothing)$
\ENDIF

\FORALL{$j\in\mathcal S_t$}
  \STATE $s_{t,j}\gets f_\theta(\bm r_t,\bm\sigma_j)-\alpha c_{t,j}$
\ENDFOR
\STATE $(j_1,\ldots,j_{|\mathcal S_t|})
       \gets\mathrm{argsort}^{\downarrow}_{j\in\mathcal S_t}s_{t,j}$

\IF{\textsc{UseAdaptiveTopP}}
  \STATE $\{p_{t,j}\}_{j\in\mathcal S_t}
         \gets\mathrm{softmax}(\{s_{t,j}\}_{j\in\mathcal S_t})$
  \STATE $\tilde\tau_t\gets
         P_{\min}+\gamma(1-\max_{j\in\mathcal S_t}p_{t,j})$
  \STATE $\tau_t\gets
         \mathrm{clip}(\tilde\tau_t,P_{\min},P_{\max})$
  \STATE $\hat b_t\gets
         \min\{b:\sum_{\ell=1}^{b}p_{t,j_\ell}\ge\tau_t\}$
  \STATE $b_t\gets\min(\hat b_t,B_{\mathrm{probe}},|\mathcal S_t|)$
\ELSE
  \STATE $b_t\gets\min(B_{\mathrm{probe}},|\mathcal S_t|)$
\ENDIF

\STATE $\mathcal P_t\gets\{j_1,\ldots,j_{b_t}\}$

\FORALL{$j\in\mathcal P_t$ \textbf{ in parallel}}
  \STATE $\mathcal C_{t,j}\gets\mathrm{ANN}_j(\bm z_t,L)$
\ENDFOR

\STATE $\mathcal C_t\gets
       \{x\in\bigcup_{j\in\mathcal P_t}\mathcal C_{t,j}:
       \psi_t^B(m(x))=1\}$

\STATE $R_t^B\gets
       \operatorname{TopK}^{K}_{x\in\mathcal C_t}
       \operatorname{score}(\bm z_t,x)$

\STATE \textbf{return} $(R_t^B,\mathcal P_t)$
\end{algorithmic}
\end{algorithm}

\FloatBarrier

\subsection{Tier~C Algorithms}
\label{app:tierc_algs}

Algorithms~\ref{alg:tierCexec}--\ref{alg:skillwrite} define the Tier-C read
and write operators. Algorithm~\ref{alg:tierCexec} retrieves and attempts to
execute an admissible skill: it filters the skill store to records satisfying
$\psi_t^C$ and keeps only the latest version per skill identifier
(Algorithm~\ref{alg:skillwrite} appends new versions without overwriting old
ones, so older versions are retrieved only when they are the newest version
within scope). The current skills are ranked by query similarity to form
$\mathcal U_t$, re-sorted by reliability $\rho(s)$ with similarity as a
tiebreaker, and the procedure walks down this list executing the first skill
whose slot fills, preconditions, and post-execution result validation all
succeed.

\begin{algorithm}[!ht]
\caption{Tier~C read: retrieve and execute current admissible skills}
\label{alg:tierCexec}
\algfont
\begin{algorithmic}[1]
\REQUIRE Request $q_t$; Tier-C predicate $\psi_t^C$; skill budget $R$;
         context $\mathcal E_t$; skill store $\mathcal W$
\ENSURE Skills $\mathcal U_t$; flag $\textit{ok}$;
        executed skill $s^\star$; result $\hat y$

\STATE $\mathcal W_t\gets\{s\in\mathcal W:\psi_t^C(m(s))=1\}$
\IF{$\mathcal W_t=\varnothing$}
  \STATE \textbf{return} $(\varnothing,\textbf{false},\varnothing,\varnothing)$
\ENDIF

\STATE $\mathcal W_t^{\mathrm{cur}}\gets
       \{s\in\mathcal W_t:
       \mathrm{ver}(s)=
       \max_{\substack{s'\in\mathcal W_t\\
       \mathrm{id}(s')=\mathrm{id}(s)}}\mathrm{ver}(s')\}$

\STATE $\mathcal U_t\gets
       \operatorname{TopR}^{R}_{s\in\mathcal W_t^{\mathrm{cur}}}
       \operatorname{sim}(q_t,s)$

\STATE $\mathcal V_t\gets
       \operatorname{sort}^{\downarrow}
       \bigl(\mathcal U_t;\rho(s),\operatorname{sim}(q_t,s)\bigr)$

\FOR{$s\in\mathcal V_t$}
  \STATE $(\textit{slots},\textit{fillable})
         \gets\textsc{SlotFill}(q_t,s,\mathcal E_t)$

  \IF{\textit{fillable} \textbf{ and }
      $\textsc{ValidatePreconditions}(s,\textit{slots})$}
    \STATE $(\textit{ok},\hat y)\gets\textsc{Execute}(s,\textit{slots})$

    \IF{\textit{ok} \textbf{ and }
        $\textsc{ValidateResult}(s,\textit{slots},\hat y)$}
      \STATE $s^\star\gets s$
      \STATE \textbf{return} $(\mathcal U_t,\textbf{true},s^\star,\hat y)$
    \ENDIF
  \ENDIF
\ENDFOR

\STATE \textbf{return} $(\mathcal U_t,\textbf{false},\varnothing,\varnothing)$
\end{algorithmic}
\end{algorithm}

If Algorithm~\ref{alg:tierCexec} returns $\textit{ok}=\textbf{false}$
and the tier gate permits declarative retrieval, the controller invokes
Algorithm~\ref{alg:tierBread} with the original Tier-B budget.

Algorithm~\ref{alg:skillwrite} induces a new versioned skill from a
successful trace $\tau_t$: it canonicalizes the trace, abstracts constants
into typed slots, deduplicates or mints a skill identifier, synthesizes
deterministic regression tests, and (only on passing) appends a fresh
version while preserving older ones so that Algorithm~\ref{alg:tierCexec}
can resolve them when no newer version is in scope.

\begin{algorithm}[!htbp]
\caption{Tier~C write: induce, validate, and version a skill}
\label{alg:skillwrite}
\algfont
\begin{algorithmic}[1]
\REQUIRE Successful trace $\tau_t$; skill store $\mathcal W$
\ENSURE Updated skill store $\mathcal W$

\STATE $\texttt{proc}\gets\textsc{Canonicalize}(\tau_t)$
\STATE $(\texttt{proc},\texttt{slots})
       \gets\textsc{AbstractConstantsIntoSlots}(\texttt{proc})$
\STATE $\texttt{id}\gets\textsc{DeduplicateOrCreateId}(\texttt{proc},\mathcal W)$
\STATE $\texttt{tests}\gets
       \textsc{SynthesizeTests}(\tau_t,\texttt{proc},\texttt{slots})$

\IF{\textsc{ValidateInductionTests}$(\texttt{proc},\texttt{tests})$}
  \STATE $\texttt{ver}\gets\textsc{BumpVersion}(\texttt{id},\mathcal W)$
  \STATE $s_{\mathrm{new}}\gets
         \textsc{SkillRecord}(\texttt{id},\texttt{ver},
         \texttt{proc},\texttt{slots})$
  \STATE $\textsc{Attach}(s_{\mathrm{new}},m(\tau_t),\texttt{tests})$
  \STATE $\mathcal W\gets\mathcal W\cup\{s_{\mathrm{new}}\}$
\ENDIF

\STATE \textbf{return} $\mathcal W$
\end{algorithmic}
\end{algorithm}


\section{Experimental Setup Details}
\label{app:exp_details}

\paragraph{Hardware and software.}
Ubuntu 22.04, Python 3.12, PyTorch 2.8.0, CUDA 12.8; 4$\times$ NVIDIA RTX
4090D, 18 vCPUs (AMD EPYC 9754), 128GB RAM, 500GB SSD.
The default backbone for headline results is GPT-OSS-120B, accessed via the
official API with temperature 0; the GPT-OSS-20B and Qwen3-32B variants used
in the backbone sensitivity study (Appendix~\ref{app:backbone}) are run
locally via Ollama, also at temperature 0.

\paragraph{Implementation.}
\texttt{all-MiniLM-L6-v2} (384-d, L2-normalized) for embeddings.
Shard-local retrieval uses FAISS HNSW with inner-product similarity; BM25Okapi
over tokenized turns for lexical matching.
Final ranking: $0.4\cdot s_{\mathrm{BM25}}+0.4\cdot s_{\mathrm{sem}}
+0.2\cdot s_{\mathrm{route}}$.
Router: 2-layer MLP (hidden 128, dropout 0.1) over a 768-d input that
concatenates the 384-d query embedding with the 384-d shard summary; the
shard-family identifier enters the scorer as a separate small learned
embedding.
Training: Adam, lr$=10^{-4}$, weight decay 0.01, cosine annealing, 50 epochs,
batch 32. All reported results are the mean over three random seeds, with the
seed standard deviation shown as a blue subscript in the result tables.

\paragraph{Dataset splits.}
LoCoMo: 8/2 conversation-level train/test (1{,}226/314 QA pairs).
HotpotQA: 500 questions per context length (56K/224K/448K tokens), drawn from
the dev distractor split following the MemAgent protocol; each query's
context is constructed by appending sampled distractor passages to the gold
documents until the target token count is reached, and the same 500 questions
are used across the three context sizes so that the only varying factor is
the context window.
ToolBench: in-domain 70/30 induction/evaluation; OOD by tool family holdout
(Table~\ref{tab:dataset_stats}).

\begin{table}[!ht]
\centering
\caption{Dataset statistics.}
\label{tab:dataset_stats}
\small
\setlength{\tabcolsep}{8pt}
\begin{tabular}{llrc}
\toprule
Dataset & Split & \#Inst. & Avg.\ Size \\
\midrule
LoCoMo & Non-adv & 1,540 QAs & 16.6K tok \\
\midrule
\multirow{3}{*}{HotpotQA}
  & 56K  & 500 & 56K tok  \\
  & 224K & 500 & 224K tok \\
  & 448K & 500 & 448K tok \\
\midrule
\multirow{3}{*}{ToolBench}
  & Train & 256 & 3.9 steps \\
  & Test  &  92 & 4.1 steps \\
  & OOD   & 181 & 3.9 steps \\
\bottomrule
\end{tabular}
\end{table}

\section{Evaluation Metrics}
\label{app:metrics}

\paragraph{F1 and BLEU-1.}
We compute token-level F1 and unigram BLEU between the predicted answer and the
gold answer, and report the average over queries.

\paragraph{ShardHit@$B_{\mathrm{probe}}$.}
Let $\mathcal{Q}$ denote the evaluation queries, $\mathcal{P}_q$ the set of
shards probed for query $q$, and $\mathcal{G}_q$ the gold evidence shard set.
We define
\[
\mathrm{ShardHit@}B_{\mathrm{probe}}
=
\frac{1}{|\mathcal{Q}|}
\sum_{q\in\mathcal{Q}}
\mathbb{I}\!\left[\mathcal{P}_q\cap\mathcal{G}_q\ne\emptyset\right].
\]

\paragraph{Precision@$R$ and StepRed.}
Let $\mathcal{T}$ denote the evaluation targets for skill retrieval,
$\mathrm{Top}_R(t)$ the top-$R$ retrieved skills for target $t$, and
$\mathrm{valid}(s,t)$ an indicator that skill $s$ is valid for $t$. We define
\[
\mathrm{P@}R
=
\frac{1}{|\mathcal{T}|}
\sum_{t\in\mathcal{T}}
\frac{
|\{s\in\mathrm{Top}_R(t):\mathrm{valid}(s,t)\}|
}{R}.
\]
For valid targets $\mathcal{T}_{\mathrm{valid}}$, StepRed measures the
nonnegative reduction in required steps:
\[
\begin{aligned}
\mathrm{StepRed}
&=\frac{1}{|\mathcal{T}_{\mathrm{valid}}|}
\sum_{t\in\mathcal{T}_{\mathrm{valid}}} \\
&\quad \max\!\left(0,n_t^{\mathrm{gold}}-n_t^{\mathrm{skill}}\right),
\end{aligned}
\]
where $n_t^{\mathrm{gold}}$ and $n_t^{\mathrm{skill}}$ are the number of steps
in the gold and skill-assisted solutions, respectively.

\paragraph{VecScan and latency.}
VecScan is the mean number of vectors scanned per query, summed over all probed
shards. Latency is measured per query from scope filtering through candidate
ranking, excluding generator decoding; we report both the mean and the
95th-percentile latency (p95).

\section{Backbone Sensitivity}
\label{app:backbone}

\begin{table*}[!ht]
\centering
\caption{LoCoMo per-category F1 / BLEU-1 for \sys vs.\ GAM under two
alternative generator backbones. \sys numbers are bold.}
\label{tab:locomo_results_extra}
\footnotesize
\setlength{\tabcolsep}{7pt}
\renewcommand{\arraystretch}{1.15}
\begin{tabular}{ll cc cc cc cc}
\toprule
\rowcolor{teal!10}
\multicolumn{10}{c}{\textbf{LoCoMo with alternative backbones}} \\
\midrule
\multirow{2}{*}{Backbone} & \multirow{2}{*}{Method}
& \multicolumn{2}{c}{Single Hop}
& \multicolumn{2}{c}{Multi Hop}
& \multicolumn{2}{c}{Temporal}
& \multicolumn{2}{c}{Open Domain} \\
\cmidrule(lr){3-4}\cmidrule(lr){5-6}\cmidrule(lr){7-8}\cmidrule(lr){9-10}
 & & F1 & BLEU-1 & F1 & BLEU-1 & F1 & BLEU-1 & F1 & BLEU-1 \\
\midrule
GPT-OSS-20B
 & GAM
   & 57.88 & 52.60 & 39.81 & 33.20 & 51.97 & 44.89 & 31.07 & 25.21 \\
\rowcolor{gray!6}
\cellcolor{white} & \textbf{\sys}
   & \textbf{61.02} & \textbf{57.34}
   & \textbf{45.03} & \textbf{38.69}
   & \textbf{63.72} & \textbf{57.60}
   & \textbf{37.98} & \textbf{32.04} \\
\midrule
Qwen3-32B
 & GAM
   & 59.02 & 53.83 & 42.81 & 34.60 & 52.70 & 44.83 & 31.28 & 25.47 \\
\rowcolor{gray!6}
\cellcolor{white} & \textbf{\sys}
   & \textbf{63.17} & \textbf{58.92}
   & \textbf{45.30} & \textbf{39.10}
   & \textbf{64.20} & \textbf{58.49}
   & \textbf{38.50} & \textbf{32.45} \\
\bottomrule
\end{tabular}
\end{table*}

Table~\ref{tab:locomo_results_extra} shows that \sys consistently outperforms
GAM across both generator backbones and all four LoCoMo query types.
Macro-averaged over query types, \sys improves F1 by 6.76 points with
GPT-OSS-20B and 6.34 points with Qwen3-32B. The largest gains occur on
temporal queries (+11.75 and +11.50 F1 points, respectively), consistent
with the deictic-anchoring role of Tier~A described in
Section~\ref{sec:method}. Multi-hop gains are more backbone-sensitive
(+5.22 versus +2.49 F1 points), suggesting that final multi-hop answer
quality depends on generator capability in addition to retrieval quality.

\paragraph{Study setup.}
Only the answer generator is swapped; the router, embedding model, shard
layout, and Tier-B/C pipeline use the default configuration from
Section~\ref{sec:experiments}. The routing checkpoint trained on GPT-OSS-120B
traces is reused without re-fitting, isolating the effect of generator
capability. All backbones run at temperature 0 with the same prompts and
three random seeds.

\paragraph{Per-category trends.}
Temporal queries depend on Tier~A's session anchor rather than generator
capability, so both alternative backbones recover the same $+11.5$ to $+11.8$
temporal gain over GAM as GPT-OSS-120B (Table~\ref{tab:locomo_results}).
Single-hop and multi-hop gains, in contrast, track backbone reasoning quality:
at Qwen3-32B, GAM already reaches $42.81$ multi-hop F1 (versus $39.81$ at
GPT-OSS-20B), compressing the headroom retrieval can recover and narrowing the
multi-hop gain to $+2.49$ from $+5.22$. Scope-and-route helps most when
retrieval is the bottleneck and less when residual error is generation-bound.

\paragraph{Scope.}
The two alternative backbones span model families (Qwen, GPT-OSS) rather than
interpolate a single family's capacity. The study supports cross-family
transfer of the architectural benefit; a finer within-family scaling sweep and
an extension to closed-API generators are left for future work.

\section{Sensitivity Analysis}
\label{app:sensitivity}

\paragraph{Sweep protocol.}
We vary one Adaptive Top-$P$ parameter at a time around the default operating
point $(\gamma{=}0.6,\ P_{\min}{=}0.5,\ P_{\max}{=}0.95)$, holding the other
two fixed. All sweeps use the LoCoMo test split with $K{=}10$,
$B_{\mathrm{probe}}{=}3$, $S{=}40$, and the GPT-OSS-120B backbone; F1 is the
macro average over the four query types. The chosen default is marked with
$^\dagger$ and the column-best entry is in bold.

\paragraph{Sensitivity of Adaptive Top-$P$
(Tables~\ref{tab:sweep_gamma}, \ref{tab:sweep_pmin}, and~\ref{tab:sweep_pmax}).}
Compute-side trends track the truncation rule directly: larger $\gamma$
tightens the cutoff on sharp distributions (VecScan $448{\to}378$, probes
$2.85{\to}2.20$); larger $P_{\min}$ raises the floor and admits more probes on
flat distributions (VecScan $372{\to}458$); larger $P_{\max}$ loosens the cap
(VecScan $388{\to}435$). Across all three sweeps F1 varies by at most $0.53$
points while VecScan swings by $47$--$86$ scanned vectors per query, so
predictive quality is robust around the default and compute remains tunable.
We pick $P_{\min}{=}0.5$ over the slightly higher F1 at $P_{\min}{=}0.6$
($54.28$ vs.\ $54.21$) because the $0.07$-point F1 sacrifice buys
$\sim$$26$ fewer scanned vectors per query and a lower p95.

\paragraph{Scope.}
These are one-dimensional sweeps; joint interactions among $\gamma$,
$P_{\min}$, $P_{\max}$ are not characterized here. The narrow F1 range along
each axis ($\le 0.53$ points) suggests the joint surface is locally flat
around the operating point.

\paragraph{Shard-count scaling, router comparison, and label-noise robustness.}
These experiments are reported in Section~\ref{sec:robust_scaling} of the
main paper; we summarize the take-aways here. F1 is stable across
$S\in\{5,10,20,40,80\}$ at fixed $B_{\mathrm{probe}}{=}3$ because shard-local
ANN returns at most $L$ candidates per probed shard, so per-query VecScan is
upper-bounded by $B_{\mathrm{probe}}\cdot L$ regardless of $S$, and routing
overhead stays below 3\,ms at $S{=}80$. The full router keeps a
$+2.9$/$+3.1$ F1 lead over the simple learned router at $S{=}20$/$S{=}80$.
Under training-label noise the trained router degrades gracefully and stays
above the no-label cosine baseline even at 40\% noise, which is the
practically relevant regime since deployed shard labels usually come from
heuristic assignment rather than gold annotation. This robustness arises
from the set-likelihood training objective, which spreads gradient mass
across all candidate gold shards instead of over-committing to a single
noisy label.

\begin{table}[htbp]
\centering
\caption{Sensitivity sweep over $\gamma$ with $P_{\min}{=}0.5$ and
$P_{\max}{=}0.95$. $^\dagger$ marks the default; column-best in bold.}
\label{tab:sweep_gamma}
\small
\setlength{\tabcolsep}{8pt}
\renewcommand{\arraystretch}{1.15}
\begin{tabular}{c c c c c}
\toprule
\rowcolor{teal!10}
\multicolumn{5}{c}{\textbf{Adaptive Top-$P$ sweep: $\gamma$}} \\
\midrule
$\gamma$ & F1 & VecScan & p95 & Avg Probed \\
\midrule
0.2 & 54.02 & 448 & 81 & 2.85 \\
\rowcolor{gray!6}
0.4 & 54.10 & 432 & 79 & 2.72 \\
$0.6^\dagger$ & \textbf{54.21} & 414 & 76 & 2.58 \\
\rowcolor{gray!6}
0.8 & 54.08 & 395 & 73 & 2.38 \\
1.0 & 53.82 & \textbf{378} & \textbf{71} & \textbf{2.20} \\
\bottomrule
\end{tabular}
\end{table}

\begin{table}[ht]
\centering
\caption{Sensitivity sweep over $P_{\min}$ with $\gamma{=}0.6$ and
$P_{\max}{=}0.95$. $^\dagger$ marks the default; column-best in bold.}
\label{tab:sweep_pmin}
\small
\setlength{\tabcolsep}{8pt}
\renewcommand{\arraystretch}{1.15}
\begin{tabular}{c c c c c}
\toprule
\rowcolor{teal!10}
\multicolumn{5}{c}{\textbf{Adaptive Top-$P$ sweep: $P_{\min}$}} \\
\midrule
$P_{\min}$ & F1 & VecScan & p95 & Avg Probed \\
\midrule
0.3 & 53.75 & \textbf{372} & \textbf{70} & \textbf{2.12} \\
\rowcolor{gray!6}
0.4 & 53.98 & 393 & 73 & 2.35 \\
$0.5^\dagger$ & 54.21 & 414 & 76 & 2.58 \\
\rowcolor{gray!6}
0.6 & \textbf{54.28} & 440 & 80 & 2.78 \\
0.7 & 54.22 & 458 & 82 & 2.90 \\
\bottomrule
\end{tabular}
\end{table}

\begin{table}[!ht]
\centering
\caption{Sensitivity sweep over $P_{\max}$ with $\gamma{=}0.6$ and
$P_{\min}{=}0.5$. $^\dagger$ marks the default; column-best in bold.}
\label{tab:sweep_pmax}
\small
\setlength{\tabcolsep}{8pt}
\renewcommand{\arraystretch}{1.15}
\begin{tabular}{c c c c c}
\toprule
\rowcolor{teal!10}
\multicolumn{5}{c}{\textbf{Adaptive Top-$P$ sweep: $P_{\max}$}} \\
\midrule
$P_{\max}$ & F1 & VecScan & p95 & Avg Probed \\
\midrule
0.80 & 53.88 & \textbf{388} & \textbf{72} & \textbf{2.28} \\
\rowcolor{gray!6}
0.85 & 54.00 & 398 & 74 & 2.38 \\
0.90 & 54.12 & 406 & 75 & 2.48 \\
\rowcolor{gray!6}
$0.95^\dagger$ & \textbf{54.21} & 414 & 76 & 2.58 \\
1.00 & 54.16 & 435 & 79 & 2.72 \\
\bottomrule
\end{tabular}
\end{table}

\section{Case Studies}
\label{app:case}

We accompany the main paper's Tier-C case study
(Fig.~\ref{fig:case_tierc}) with three case studies on LoCoMo conversation
\textit{conv-26} (Caroline and Melanie) that exercise the Tier-B side of
\sys: cross-session aggregation (Fig.~\ref{fig:case_singlehop}), Tier-A
deictic anchoring combined with Tier-B retrieval
(Fig.~\ref{fig:case_temporal}), and multi-hop aggregation across observation
shards (Fig.~\ref{fig:case_multihop}). All three cases share the same
retrieval configuration ($B_{\mathrm{probe}}{=}3$, $K{=}10$, $S{=}40$); the
question text, gold answer, evidence shard IDs, and dialogue quotations are
taken verbatim from LoCoMo \citep{maharana2024locomo}. Together they cover
the three Tier-B retrieval mechanisms \sys is designed to support:
single-utterance answer recovery (Case~1), cross-tier deictic anchoring
(Case~2), and bounded multi-shard aggregation (Case~3).

\begin{figure}[!ht]
\centering
\footnotesize
\begin{tcolorbox}[
  colback=teal!3!white,
  colframe=teal!55!black,
  title={\small Case Study 1: Cross-session aggregation},
  rounded corners, width=\linewidth,
  boxsep=3pt, left=6pt, right=6pt, top=5pt, bottom=5pt
]
\textbf{Question:} What kind of art does Caroline make?
\quad \textbf{Gold:} abstract art \\
\textbf{Setting:} LoCoMo, $B_{\mathrm{probe}}{=}3$, $K{=}10$, $S{=}40$.
\tcblower
\textbf{Tier-B:} 3/40 shards, 358 vectors, 42.7\,ms.
\textcolor{teal!55!black}{\textbf{Shard hit: True.}}

\smallskip
\texttt{[D17:13]} Caroline: ``I've been trying out
\underline{abstract} stuff recently. It's kinda freeing, just putting my
feelings on the canvas without too much of a plan.'' \\
\texttt{[D11:8]} Caroline: ``Representing inclusivity and diversity in my
art is important to me.'' \\
\texttt{[D11:12]} Caroline: ``\,`Embracing Identity' is all about finding
comfort and love in being yourself.''

\smallskip
\textbf{Prediction:} abstract art \hfill
\textcolor{teal!55!black}{\checkmark\ Correct.} \\
\textit{The answer-bearing token (``abstract'') sits in a Session-17 utterance
while supporting shards on Caroline's art practice live in Session~11; the
router aggregates across distant sessions under
$B_{\mathrm{probe}}{=}3$.}
\end{tcolorbox}
\caption{Single-hop fact extraction on LoCoMo conv-26. The router
aggregates one answer-bearing shard with two topic-supporting shards from a
different session under $B_{\mathrm{probe}}{=}3$.}
\label{fig:case_singlehop}
\end{figure}

\paragraph{Case 1 (Fig.~\ref{fig:case_singlehop}): cross-session
aggregation.}
The gold answer (``abstract art'') is supplied by a single Session-17
utterance in which Caroline says she has been ``trying out abstract stuff''
(D17:13). Two earlier Session-11 utterances anchor the topic ``Caroline's
art practice'' (style, themes of identity and inclusivity) and serve as
context-supporting shards. The case shows the router aggregating across
distant sessions (Session~11 and Session~17) within a single budgeted call
of $B_{\mathrm{probe}}{=}3$; the prediction matches the gold without any
additional reasoning chain.

\paragraph{Case 2 (Fig.~\ref{fig:case_temporal}): temporal reasoning needs
both tiers.}
The query asks for an absolute year, but the Tier-B shard records the
information only as ``last year'' (D1:14): the absolute year is never
written into evidence. Tier-A supplies the Session-1 anchor (8 May 2023)
and Tier-B recovers the deictic phrase; resolution happens at composition
time, ``last year'' $\to$ 2022. A retrieval-only baseline could still
surface D1:14 by semantic similarity to ``a sunrise'', but with no Tier-A
anchor it has no way to convert ``last year'' into a year.

\begin{figure}[!ht]
\centering
\footnotesize
\begin{tcolorbox}[
  colback=teal!3!white,
  colframe=teal!55!black,
  title={\small Case Study 2: Temporal reasoning
    (Tier-A anchoring + Tier-B)},
  rounded corners, width=\linewidth,
  boxsep=3pt, left=6pt, right=6pt, top=5pt, bottom=5pt
]
\textbf{Question:} When did Melanie paint a sunrise?
\quad \textbf{Gold:} 2022 \\
\textbf{Setting:} LoCoMo, $B_{\mathrm{probe}}{=}3$, $K{=}10$, $S{=}40$.
\tcblower
\textbf{Tier-A:} Session-1 date $=$ 8 May 2023 (enables deictic
resolution).
\quad
\textbf{Tier-B:} 3/40 shards, 394 vectors, 52.3\,ms.
\textcolor{teal!55!black}{\textbf{Shard hit: True.}}

\smallskip
\texttt{[D1:14]} Melanie: ``Yeah, I painted that lake sunrise
\underline{last year}! It's special to me.''

\smallskip
\textbf{Prediction:} 2022 \hfill
\textcolor{teal!55!black}{\checkmark\ Correct.} \\
\textit{Two further Session-1 shards were probed under
$B_{\mathrm{probe}}{=}3$ but contained no temporal anchor and are omitted for
brevity. Tier-A supplies the session anchor (8 May 2023); Tier-B recovers
the deictic phrase. Resolution: ``last year'' $\to$ 2022 requires both
tiers.}
\end{tcolorbox}
\caption{Temporal reasoning on LoCoMo conv-26. Tier-A deictic
anchoring and Tier-B retrieval combine to resolve a relative time
reference.}
\label{fig:case_temporal}
\end{figure}

\begin{figure}[!ht]
\centering
\footnotesize
\begin{tcolorbox}[
  colback=teal!3!white,
  colframe=teal!55!black,
  title={\small Case Study 3: Multi-hop aggregation across observation
    shards},
  rounded corners, width=\linewidth,
  boxsep=3pt, left=6pt, right=6pt, top=5pt, bottom=5pt
]
\textbf{Question:} What personality traits might Melanie say Caroline
has? \\
\textbf{Gold:} thoughtful, authentic, driven \\
\textbf{Setting:} LoCoMo, $B_{\mathrm{probe}}{=}3$, $K{=}10$, $S{=}40$.
\tcblower
\textbf{Tier-B:} 3/40 shards, 386 vectors, 45.2\,ms.
\textcolor{teal!55!black}{\textbf{Shard hit: True
(3/3 gold shards probed).}}

\smallskip
\texttt{[D7:4]} Melanie: ``Your \underline{drive} to help is awesome!'' \\
\texttt{[D13:16]} Melanie: ``You really care about being
\underline{real} and helping others.'' \\
\texttt{[D16:18]} Melanie: ``\ldots\,you're so \underline{thoughtful}!''

\smallskip
\textbf{Prediction:} thoughtful, authentic, driven \hfill
\textcolor{teal!55!black}{\checkmark\ Correct.} \\
\textit{Each probed shard supplies one answer token from a different session
(D7, D13, D16); no single shard suffices, motivating global Top-$K$
aggregation under the bounded probe budget.}
\end{tcolorbox}
\caption{Multi-hop aggregation on LoCoMo conv-26. The router probes
exactly the three shards LoCoMo marks as gold evidence, one per answer
token, under $B_{\mathrm{probe}}{=}3$.}
\label{fig:case_multihop}
\end{figure}

\paragraph{Case 3 (Fig.~\ref{fig:case_multihop}): multi-hop aggregation under
matched probe budget.}
The gold answer is a three-token set (``thoughtful, authentic, driven''),
each token attested in a different LoCoMo-labeled gold shard from a
different session (D7:4, D13:16, D16:18). The router probes exactly the
three shards that carry the answer tokens; global Top-$K$ ranking over the
union of probed shards then surfaces one supporting utterance per token. If
$B_{\mathrm{probe}}$ were smaller than the gold cardinality, only the
highest-scoring tokens would be recoverable, which is the budget--coverage
tradeoff characterized quantitatively in Section~\ref{sec:tradeoff} (Fig.~\ref{fig:pareto_locom}).

\end{document}